\documentclass[11pt,a4paper]{article}
\usepackage[hyperref]{emnlp-ijcnlp-2019}
\usepackage{times}
\usepackage{latexsym}
\usepackage{multirow}
\usepackage{graphicx}
\usepackage{CJKutf8}

\usepackage{url}

\aclfinalcopy 

\newcommand*{\enja}{EN~$\rightarrow$~JA}
\newcommand*{\jaen}{JA~$\rightarrow$~EN}

\title{Designing the Business Conversation Corpus}

\author{
    Mat\={\i}ss Rikters\thanks{\ \ equal contribution} \and Ryokan Ri\footnotemark[1] \and Tong Li\footnotemark[1] \and Toshiaki Nakazawa \\
    The University of Tokyo \\
    7-3-1 Hongo, Bunkyo-ku, Tokyo, Japan \\
    {\tt \{matiss, li0123, litong, nakazawa\}@logos.t.u-tokyo.ac.jp} \\
  }

\date{}

\begin{document}
\maketitle
\begin{abstract}
    While the progress of machine translation of written text has come far in the past several years thanks to the increasing availability of parallel corpora and corpora-based training technologies, automatic translation of spoken text and dialogues remains challenging even for modern systems. In this paper, we aim to boost the machine translation quality of conversational texts by introducing a newly constructed Japanese-English business conversation parallel corpus. A detailed analysis of the corpus is provided along with challenging examples for automatic translation. We also experiment with adding the corpus in a machine translation training scenario and show how the resulting system benefits from its use.
\end{abstract}

\section{Introduction}
\label{sec:intro}

There are a lot of ready-to-use parallel corpora for training machine translation systems, however, most of them are in written languages such as web crawl, news-commentary\footnote{http://www.statmt.org/wmt19/translation-task.html}, patents \cite{ntcir9-patmt}, scientific papers \cite{NAKAZAWA16.621} and so on. Even though some of the parallel corpora are in spoken language, they are mostly spoken by only one person (in other words, they are monologues) \cite{cettoloEtAl:EAMT2012,mustc19} or contain a lot of noise \cite{tiedemann-2016-finding,pryzant_jesc_2018}. Most of the machine translation evaluation campaigns such as WMT\footnote{http://www.statmt.org/wmt19/}, IWSLT\footnote{http://workshop2019.iwslt.org} and WAT\footnote{http://lotus.kuee.kyoto-u.ac.jp/WAT/} adopt the written language, monologue or noisy dialogue parallel corpora for their translation tasks. Among them, there is only one clean, dialogue parallel corpus \cite{salesky2018slt} adopted by IWSLT in the conversational speech translation task, nevertheless, the availability of such kind of corpus is still limited.

The quality of machine translation for written text and monologue has vastly improved due to the increase in the amount of the available parallel corpora and the recent neural network technologies. However, there is much room for improvement in the context of dialogue or conversation translation. One typical case is the translation from pro-drop language to the non-pro-drop language where correct pronouns must be supplemented according to the context. The omission of the pronouns occurs more frequently in spoken language than written language. Recently, context-aware translation models attract attention from many researchers \cite{tiedemann-scherrer-2017-neural,voita-etal-2018-context,voita-etal-2019-good} to solve this kind of problem, however, there are almost no conversational parallel corpora with context information except noisy OpenSubtitles corpus.

\begin{CJK}{UTF8}{min}
\begin{table*}[t]
\centering
\begin{small}
\begin{tabular}{|@{ }l@{ }|@{ }p{6cm}@{ }|@{ }l@{ }|@{ }p{6cm}@{ }|}
 \multicolumn{4}{c}{Scene: telephone consultation about intrafirm export} \\ \hline
 \multicolumn{2}{|c|}{Japanese} & \multicolumn{2}{|c|}{English} \\ \hline 
 Speaker & Content & Speaker & Content \\ \hline
 山本 & もしもし、山本と申します。 & Yamamoto & Hello, this is Yamamoto. \\ \hline
 田中 & 販売部門の田中と申します。 & Tanaka & This is Tanaka from the Department of Sales. \\ \hline
 田中 & 輸出に関してご助言いただきたくお電話しました。 & Tanaka & I called you to get some advice from you concerning export. \\ \hline
 山本 & はい、どのようなご用件でしょう？ & Yamamoto & Okay, what's the matter? \\ \hline
 田中 & イランの会社から遠視カメラの引き合いを受けているのですが、イランに対しては輸出制限があると新聞で読んだことがある気がして。 & Tanaka & We got an inquiry from an Iranian company about our far-sight cameras, but I think I read in the newspaper that there are export restrictions against Iran. \\ \hline
 田中 & うちで売っているようなカメラなら、特に問題にならないのでしょうか？ & Tanaka & Is there no problem with cameras like the ones we sell? \\ \hline
 山本 & 恐れ入りますが、イランへの輸出は、かなり制限されているのが事実です。 & Yamamoto & I'm afraid that the fact is, exports to Iran are highly restricted. \\ \hline
 ... & ... & ... & ... \\ \hline
\end{tabular}
\end{small}
\caption{An example of the Japanese-English business conversation parallel corpus.}
\label{table:corpus-example}
\end{table*}
\end{CJK}

Taking into consideration the factors mentioned above, a document and sentence-aligned conversational parallel corpus should be advantageous to push machine translation research in this field to the next stage. In this paper, we introduce a newly constructed Japanese-English business conversation parallel corpus. This corpus contains 955 scenarios, 30,000 parallel sentences. Table \ref{table:corpus-example} shows an example of the corpus. 

An updated version of the corpus is available on GitHub\footnote{https://github.com/tsuruoka-lab/BSD} under the Creative Commons Attribution-NonCommercial-ShareAlike (CC BY-NC-SA) license. The release contains a 20,000 parallel sentence training data set, and development and evaluation data sets of 2051 and 2120 parallel sentences respectively.

We choose the business conversation as the domain of the corpus because 1) the business domain is neither too specific nor too general, and 2) we think that a clean conversational parallel corpus is useful to open new machine translation research directions. We hope that this corpus becomes one of the standard benchmark data sets for machine translation. 

\begin{CJK}{UTF8}{min}
What is unique for this corpus is that each scenario is annotated with scene information, as shown in the top of Table \ref{table:corpus-example}. In conversations, the utterances are often very short and vague, therefore it is possible that they should be translated differently depending on the situations where the conversations are taking place. For example, Japanese expression 「すみません」 can be translated into several English expressions such as ``Excuse me.'' (when you call a store attendant), ``Thank you.'' (when you are given some gifts) or ``I'm sorry.'' (when you need to apologise). By using the scene information, it is possible to discriminate the translations, which is hard to do with only the contextual sentences. Furthermore, it might be possible to connect the scene information to the multi-modal translation, which is also hardly studied recently, such as estimating the scenes by the visual information.
\end{CJK}

The structure of this paper is as follows: we explain how the corpus is constructed in Section \ref{sec:statistics}, show the fundamental analysis of the corpus in Section \ref{sec:error-analysis}, report results of machine translation experiments in Section \ref{sec:mt-exp}, and give a conclusion in Section \ref{sec:conclusion}.

\section{Description and Statistics of the Corpus}
\label{sec:statistics}

\begin{table}[t]
    \centering
    \begin{small}
    \begin{tabular}{|c|c|c|} 
         \hline
         Scene & Scenarios & Sentences \\ 
         \hline \hline
         \multicolumn{3}{|c|}{\jaen} \\ \hline
         face-to-face       & 165   & 5,068 \\ 
         phone call         & 77    & 2,329 \\
         general chatting   & 101   & 3,321 \\
         meeting            & 106   & 3,561 \\
         training           & 16    & 608 \\
         presentation       & 4     & 113 \\ 
         \hline
         sum                & 469   & 15,000 \\
         \hline \hline
         \multicolumn{3}{|c|}{\enja} \\ \hline
         face-to-face       & 158   & 4,876 \\ 
         phone call         & 99    & 2,949 \\
         general chatting   & 102   & 2,988 \\
         meeting            & 103   & 3,315 \\
         training           & 9     & 326 \\
         presentation       & 15    & 546 \\
         \hline
         sum                & 486   & 15,000 \\
         \hline
    \end{tabular}
    \end{small}
    \caption{Statistics for the corpus, where \jaen\ represents scenarios which are written in Japanese then translated into English
    and \enja\ represents scenarios constructed in the reverse way.}
    \label{table:overview}
\end{table}

The Japanese-English business conversation corpus, namely Business Scene Dialogue (BSD) corpus, is constructed in 3 steps: 1) selecting business scenes, 2) writing monolingual conversation scenarios according to the selected scenes, and 3) translating the scenarios into the other language. The whole construction process was supervised by a person who satisfies the following conditions to guarantee the conversations to be natural:
\begin{itemize}
    \item has the experience of being engaged in language learning programs, especially for business conversations
    \item is able to smoothly communicate with others in various business scenes both in Japanese and English
    \item has the experience of being involved in business
\end{itemize}

\subsection{Business Scene Selection}

The business scenes were carefully selected to cover a variety of business situations, including meetings and negotiations, as well as so-called water-cooler chats. Details are shown in Table \ref{table:overview}. We also paid attention not to select specialised scenes which are suitable only for a limited number of industries. We made sure that all of the selected scenes are generic to a broad range of industries.

\subsection{Monolingual Dialogue Scenario Writing}

Dialogue scenarios were monolingually written for each of the selected business scenes. Half of the monolingual scenarios were written in Japanese and the other half were written in English (15,000 sentences for each language). This is because we want to cover a wide range of lexicons and expressions for both languages in the corpus. Writing the scenarios only in one language might fail to cover useful, important expressions in the other language when they are translated in the following step.

\subsection{Scenario Translation}

The monolingual scenarios were translated into the other language by human translators. They were asked to make the translations not only accurate, but also as fluent and natural as a real dialogue at the same time. This principle is adopted to eliminate several common tendencies of human translators when performing Japanese-English translation on a written text. For example, Japanese pronouns are usually omitted in a dialogue, however, when the English sentences are literally translated into Japanese, the translators tend to include unnecessary pronouns. It is acceptable as a written text, but would be rather unusual as a spoken text.

\section{Analysis of the Corpus}
\label{sec:error-analysis}

To understand the difficulty of translating conversations, we conduct an analysis regarding the newly constructed corpus. We choose to use Google Translate \footnote{https://translate.google.com/ (May 2019)}, one of the most powerful neural machine translation (NMT) systems which are publicly available, to produce the translations.

Our primary focus is to understand how many sentences require context to be properly translated. We randomly sample 10 scenarios (322 sentences) from the corpus, and check the translations for fatal translation errors, ignoring fluency or minor grammatical mistakes. As a result, 12 sentences have errors due to phrase ambiguity that needs understanding the context, or the real-world situation, and 18 errors of pronouns due to zero anaphora, which is described in the following section, in the source language (Japanese). Now we focus on the latter errors.

\subsection{Zero Anaphora}
\label{sec:anaphora}
\begin{CJK}{UTF8}{min}
As an important preliminary, we briefly introduce a grammatical phenomenon called {\it zero anaphora}. In Japanese, some arguments of verbs are often omitted from the phrases when they are obvious from the context. When translating them into English, one often has to identify the referent of the omitted argument and recover it in English, as English does not allow omitting the core arguments (i.e., subject, object). In the following Japanese example, the subject of the verb 買った is omitted, but in the English translation a pronoun, for example {\it he}, has to be recovered. Note that the subject could be anyone, not necessarily {\it he}, depending on the context. The task of identifying the referent of zero anaphora is called {\it zero anaphora resolution}, which is one of the most difficult tasks of NLP.

    \begin{table}[h]
    \begin{small}
      \begin{tabular}{llll}
        太郎は  & 買った& 牛乳を & 飲んだ \\
        Taro-SBJ & buy-PST & milk-OBJ & drink-PST \\
        \multicolumn{4}{l}{``Taro drank the milk {\bf he} bought.''}
      \end{tabular}
    \end{small}
    \end{table}

\end{CJK}

\subsection{Quantitative Analysis}
\label{sec:quantitative-anaphora}

\begin{CJK}{UTF8}{min}
To estimate how many sentences need zero anaphora resolution in the business conversation corpus, we counted the number of sentences with the personal pronouns ({\it e.g.}, 彼, 彼女, 私, あなた in Japanese, {\it I}, {\it you}, {\it he}, {\it she} in English) in both Japanese and English. As a result, 62\% of English sentences contain personal pronouns, while only 11\% of Japanese sentences do. This means about 50\% of the sentences in the corpus potentially need zero reference resolution when we translate them from Japanese into English.
\end{CJK}

\begin{figure}[t]
    \centering
    \includegraphics[width=7.5cm]{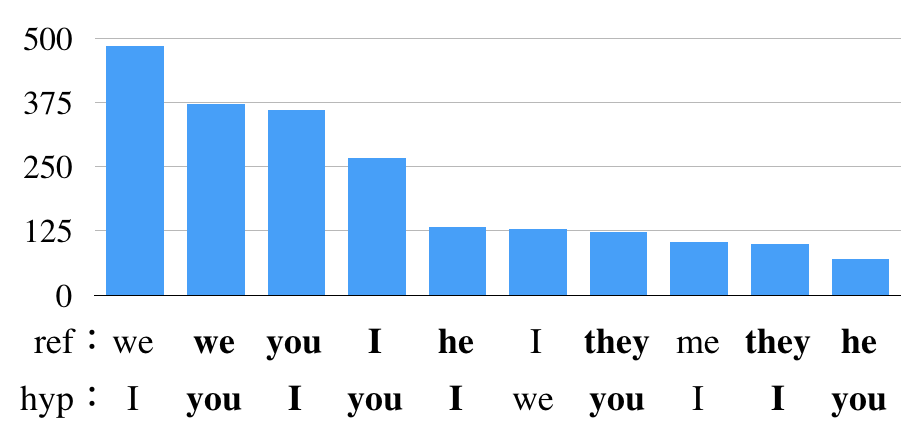}
    \caption{The top 10 frequent errors of pronoun translation (fatal errors denoted in boldface)}
    \label{fig:error-graph}
\end{figure}

\begin{CJK}{UTF8}{min}
\begin{figure*}[t]
\begin{small}
  \begin{tabular}{lp{12cm}}
   \bf Previous Source: & 支店長はポールをクビにするみたいだよ。 \\
   \bf Previous Reference: & It seems like the branch manager will be firing {\bf Paul}. \\\\

   \bf Source: & 仕事もあまりしない上に、休み、早退ばかりを希望するから。 \\
   \bf Reference: & {\bf He} doesn't work much , and {\bf he} takes days off and asks to leave early often. \\
   \bf Google Translate: & {\bf I} do not have much work , and {\bf I} would like to leave early and leave early.\\

  \end{tabular}
  \caption{An example of Japanese to English Google Translate output. The words in boldface are supposed to denote the same referent(Paul).}

  \label{fig:google-error-example1}
  \end{small}
\end{figure*}
\end{CJK}

\begin{CJK}{UTF8}{min}
\begin{figure*}[t]
\begin{small}
  \begin{tabular}{lp{12.5cm}}
   \bf Previous Source: & {\bf [Speaker1]} 彼の代わりに、優秀な人が入ってくれれば、僕の仕事量が減るはずなんだ。 \\
   \bf Previous Reference: & I think I can work less if there's someone excellent coming in as a replacement for him. \\\\

   \bf Source: & {\bf [Speaker2]} もう少しの辛抱だよ。 \\
   \bf Reference: & {\bf You} just need a bit more patience. \\
   \bf Google Translate: & {\bf I} have a little more patience.\\
  \end{tabular}
  \caption{An example of Japanese to English Google Translate output. Correct translation needs the speaker information.}

  \label{fig:google-error-example2}
\end{small}
\end{figure*}
\end{CJK}

To reveal what kinds of zero pronouns are hard to translate, we again heuristically count the number of the translation errors of the pronouns for the entire corpus. We counted the number of the translated sentences that have pronouns different from their reference sentences. By this heuristic, we detected 3,653 errors (12\% of the whole corpus). The top 10 frequent errors are shown in Figure \ref{fig:error-graph}.

Some errors such as {\it we} $\rightarrow$ {\it I}, {\it I} $\rightarrow$ {\it me}, might be not fatal, and not be regarded as translation errors. However, there are still many fatal errors among first, second and third-person pronouns (denoted in boldface in the graph).

Looking at the pronouns that the NMT system produced, we can see the tendency of the system to generate frequent pronouns such as {\it you}, {\it I}. This suggests that the current system tries to compensate source (Japanese) zero pronouns simply by generating frequent target (English) pronouns. When the referent is denoted in relatively infrequent pronouns in the target language, it is hard to be correctly translated. To deal with this problem, We need to develop more sophisticated systems that take context into account.

\subsection{Qualitative Analysis}
\label{sec:qualitative-anaphora}

This section exemplifies some zero-anaphora translation errors and discusses what kind of information is needed to perform correct translation.

\begin{CJK}{UTF8}{min}

\subsubsection*{A translation that needs world knowledge and inference}
In Figure \ref{fig:google-error-example1}, the subjects of the verbs are omitted in the source sentence 「（彼は: he）仕事もあまりしない上に、（彼は: he）休み、早退ばかりを希望するから」. This causes the NMT system to incorrectly translate the zero pronouns into {\it I}, although they actually refer to {\it Paul} in the previous sentence and thus have to be translated into {\it he}.

Resolving these zero pronouns, however, is not straightforward, even if one has access to the information of the previous sentence. For example, to identify the subject of 「仕事もあまりしない」(doesn't work much), one has to know ``laziness can lead to being fired'' and thereby infer that Paul, who is about to be fired, is the subject. Existing contextual NMT systems \citep{voita-etal-2018-context, EvaluatingDiscourse:2018vp, Maruf:2019vc} still do not seem to be able to handle this complexity.
\end{CJK}

\begin{CJK}{UTF8}{min}

\subsubsection*{A translation that needs to know who is talking}
In Figure \ref{fig:google-error-example2}, again, the subject is omitted in the source sentence 「（君は: you）もう少しの辛抱だよ。」. The NMT system incorrectly translates the zero pronouns into {\it I}.

It is worth noting that the type of the zero pronoun differs from the one in Figure \ref{fig:google-error-example1} in that the referent in Figure \ref{fig:google-error-example2} does not linguistically appear within the text (called {\it exophora}), while the other does ({\it endophora}) \citep{Brown:1983wy}. The referent of the zero pronoun in Figure \ref{fig:google-error-example2} is the listener of the utterance ({\it you}), and it usually does not have another linguistic item (such as the name of the person) that can be referred to. Although some modality expressions and verb types can give constraints to the possible referents \citep{Nakaiwa:1996vb}, essentially, the resolution of exophora needs the reference to situation.

In this case, the correct translation depends on who is speaking. In the original conversation, the utterance is from Speaker 2 to Speaker 1, and given the context, one can infer that Speaker 2 is speaking to give a consolation to Speaker 1 and thus the subject should be {\it you} (Speaker 1). However, if the utterance was from Speaker 1, he would then just be complaining about his situation saying {\it ``{\bf I} just need a bit more patience''}. This example emphasises that the speaker information is essential to translate some utterances in conversation correctly.

\end{CJK}

\section{Machine Translation Experiments}
\label{sec:mt-exp}

The BSD corpus was created with the intended use of training NMT systems. Thus, we trained NMT models using the corpus in both translation directions. As the BSD corpus is rather small for training reasonable MT systems, we also experimented with combining it with two larger conversational domain corpora. We employed translators to translate the AMI Meeting Corpus \cite{mccowan2005ami} (AMI) and the English part of Onto Notes 5.0 \cite{weischedel2013ontonotes} (ON) into Japanese with the same instructions as for translating the BSD corpus. Afterwards, we used them as additional parallel corpora in our experiments.

\subsection{Data Preparation}
\label{sec:mt-data}

Before training, we split each of the corpora into 3 parts - training, development and evaluation data sets. The sizes of each corpus are shown in Table \ref{tab:training-data-table}. We used Sentencepiece \cite{kudo2018sentencepiece} to create a shared vocabulary of 4000 tokens. We did not perform other tokenisation or truecasing for the training data. We used Mecab \cite{kudo2006mecab} to tokenise the Japanese side of the evaluation data, which we used only for scoring. The English side remained as-is.

\begin{table}[t]
\begin{small}
    \centering
    \begin{tabular}{|l|l|l|r|}
    \hline
    Data Set & Devel & Eval & Train \\ \hline
    BSD     & 1000 & 1000 & 28,000 \\
    AMI     & 1000 & 1000 & 108,499 \\ 
    ON      & 1000 & 1000 & 26,439 \\ \hline
    Total   & \multicolumn{3}{r|}{162,938} \\ \hline
    \end{tabular}
    \caption{Training, development and evaluation data statistics.}
    \label{tab:training-data-table}
    \end{small}
\end{table}

\subsection{Experiment Setup}
\label{sec:mt-setup}

We used Sockeye \cite{Sockeye:17} to train transformer architecture models with 6 encoder and decoder layers, 8 transformer attention heads per layer, word embeddings and hidden layers of size 512, dropout of 0.2, maximum sentence length of 128 symbols, and a batch size of 1024 words, checkpoint frequency of 4000 updates. All models were trained until they reached convergence (no improvement for 10 checkpoints) on development data.

For contrast we also trained statistical MT (SMT) systems using using the Moses \cite{Koehn2007Moses:Translation} toolkit and the following parameters: Word alignment using fast-align \cite{dyer2013simple}; 7-gram  translation  models  and  the `wbe-msd-bidirectional-fe-allff` reordering models; Language model trained with KenLM \cite{heafield2011kenlm}; Tuned using the improved MERT \cite{bertoldi2009improved}.

\subsection{Results}
\label{sec:mt-results}

Since there are almost no spaces in the Japanese raw texts, we used Mecab to tokenise the Japanese translations and references for scoring. The results in BLEU scores \cite{Papineni2001BLEU} are shown in Table \ref{tab:nmt-result-table-2} along with several ablation experiments on training NMT and SMT systems using only the BSD data, all 3 conversational corpora, and excluding the BSD corpus from the training data. The results show that adding the BSD to the two larger corpora significantly improves both SMT and NMT performance. For Japanese $\rightarrow$ English using only BSD as training data achieves a higher BLEU score than using only AMI and ON, while for English $\rightarrow$ Japanese the opposite is true. Nevertheless, in both translation directions using all 3 corpora outperforms the other results. 

\begin{table}[t]
\begin{small}
    \centering
    \begin{tabular}{|c|c|c|c|}
    \hline
                        \multicolumn{2}{|c|}{} & JA-EN & EN-JA \\ \hline
    \multirow{2}{*}{\textbf{BSD}}          & SMT & 1.90 & 5.16 \\ 
                                           & NMT & 8.32 & 8.34 \\ \hline
    \multirow{2}{*}{\textbf{AMI, BSD, ON}} & SMT & 7.27 & 5.76 \\ 
                                           & NMT & 12.88 & 13.53 \\ \hline
    \multirow{2}{*}{\textbf{AMI, ON}}      & SMT & 2.18 & 5.74 \\ 
                                           & NMT & 7.08 & 10.00 \\ \hline
    \end{tabular}
    \caption{NMT and SMT experiments using the conversational corpora. Evaluated on the Business Conversation evaluation set.}
    \label{tab:nmt-result-table-2}
    \end{small}
\end{table}

\begin{table}[t]
\begin{small}
    \centering
    \begin{tabular}{|l|c|c|}
    \hline
          & BLEU    & ChrF2 \\ \hline
          \multicolumn{3}{|c|}{\bf ON} \\ \hline
    \jaen & 9.08    & 34.38 \\ \hline
    \enja & 14.52   & 19.73 \\ \hline
          \multicolumn{3}{|c|}{\bf AMI} \\ \hline
    \jaen & 20.88   & 46.93 \\ \hline
    \enja & 23.35   & 30.25 \\ \hline
          \multicolumn{3}{|c|}{\bf BSD} \\ \hline
    \jaen & 12.88   & 35.37 \\ \hline
    \enja & 13.53   & 21.97 \\ \hline
    \end{tabular}
    \caption{BLEU and ChrF2 scores for all three evaluation data sets using the NMT system trained on all data.}
    \label{tab:nmt-result-table}
    \end{small}
\end{table}

\begin{CJK}{UTF8}{min}
\begin{figure*}[t]
\begin{small}
  \begin{tabular}{lp{12cm}}
   \bf Source: & では、終了する前に、この健康とストレスに関するセルフチェックシートに記入をして頂きたいと思います。 \\
   \bf Our Best NMT: & So before we finish, I'd like to fill in the health check-streams with this health and staff check-book.\\
   \bf Google Translate: & I would like you to fill out this health and stress self-check sheet before you finish.\\
   \bf Reference: & Before we finish off, we would like you to fill out this self-check sheet about health and stress. \\
  \end{tabular}
  \caption{An example of Japanese to English NMT output comparing our best NMT to Google Translate.}
  \label{fig:nmt-output-example}
\end{small}
\end{figure*}
\end{CJK}

\begin{CJK}{UTF8}{min}
\begin{small}
\begin{figure*}[t]
  \begin{tabular}{lp{12cm}}
   \bf Previous Source: & あぁ、マネージャーはよく休みを取ってるみたいですよ。 \\
   \bf Previous Reference: & Well, seems like our manager is taking quite a bit of time off. \\\\
   \bf Source: & 自分が取らないと、他の人が取らないだろと思ってるんでしょうね。 \\
   \bf Our Best NMT: & You think other people won't take it if they don't. \\
   \bf Google Translate: & If you don't take it, you might think that other people won't take it. \\
   \bf Reference: & Maybe he thinks if he doesn't take any, then nobody else will. \\
  \end{tabular}
  \caption{An example of Japanese to English NMT output, where a context-aware system could be more useful.}
  \label{fig:nmt-output-example-2}
\end{figure*}
\end{small}
\end{CJK}

We also evaluate the highest-scoring NMT system (trained on all corpora) on all 3 evaluation sets and report BLEU scores and ChrF2 scores \cite{popovic2015chrf} in Table \ref{tab:nmt-result-table}. We do this to verify that the models are not overfitting on the BSD data, i.e. BLEU and ChrF2 scores are not significantly higher for the BSD evaluation sets when compared to the ON and AMI sets. Results on the ON evaluation set are fairly similar to the BSD results, while results on the AMI evaluation set are noticeably higher. This can be explained by the fact that the AMI training data set is approximately four times larger than the BSD training data set, and the ON training data set is about the same size as the BSD set.

\subsection{Machine Translation Examples}

In Figure \ref{fig:nmt-output-example} we can see one of the difficult situations mentioned in Section \ref{sec:qualitative-anaphora}, where MT systems find it challenging to generate the correct pronouns in the translation. Of the three pronouns that are in the reference (we, we, you), each system translates one correctly and fails to translate the rest - both systems generate \textit{I} where it should have been \textit{we}, but our system completely omits \textit{you} while Google Translate generates \textit{you} where it should have been \textit{we}.

Figure \ref{fig:nmt-output-example-2} shows an example where both - our translation and the one from Google Translate are acceptable at the sentence-level, but when looking at the previous source and reference it becomes clear that different personal pronouns should have been used. Our system did generate ``they" in the second part of the sentence, which could be a more casual alternative to ``he", but both systems still failed to find the correct pronoun for the first part by producing ``you" instead of ``he". This is an issue that can not be fully resolved by using sentence-level MT and requires a document-level or context-aware solutions.

\section{Conclusion}
\label{sec:conclusion}
In this paper, we presented a parallel corpus of English-Japanese business conversations. The intended use-cases for the corpus are machine translation system training and evaluation. We describe the corpus in detail and indicate which linguistic phenomena are challenging to translate even for modern MT systems. We also show how adding the BSD corpus to machine translation system training helps to improve translation output of conversational texts.

We point out several examples, where sentence-level MT is unable to produce the correct translation due to lack of context from previous sentences. As the corpus is both - sentence-aligned and document-aligned, we hope that it gets used and inspires new future work such directions as document-level and context-aware neural machine translation, as well as analysing other linguistic phenomena that are relevant to translating conversational texts.

In the near future, we plan to release the full set of business conversational corpora. The set will contain all 3 corpora described in section \ref{sec:mt-exp} - an extended version of the Business Scene Dialogue corpus as well as parallel versions of the AMI Meeting Corpus and Onto Notes 5.0.

\section*{Acknowledgements}

This work was supported by “Research and Development of Deep Learning Technology for Advanced Multilingual Speech Translation”, the Commissioned Research of National Institute of Information and Communications Technology (NICT), JAPAN.

\bibliography{emnlp-ijcnlp-2019}
\bibliographystyle{acl_natbib}

\end{document}